# An automated approach for task evaluation using EEG signals

Vishal Anand[1*], S. R. Sreeja[2], Debasis Samanta[2],

[1] Department of Mechanical Engineering
Indian Institute of Technology Kharagpur, Kharagpur India
*vishal.anand456@iitkgp.ac.in
[2] Department of Computer Science and Engineering,
Indian Institute of Technology Kharagpur, Kharagpur, India
sreejasr@iitkgp.ac.in, dsamanta@iitkgp.ac.in,

**Abstract.** Critical task and cognition-based environments, such as in military and defense operations, aviation user-technology interaction evaluation on UI, understanding intuitiveness of a hardware model or software toolkit, etc. require an assessment of how much a particular task is generating mental workload on a user. This is necessary for understanding how those tasks, operations, and activities can be improvised and made better suited for the users so that they reduce the mental workload on the individual and the operators can use them with ease and less difficulty. However, a particular task can be gauged by a user as simple while for others it may be difficult. Understanding the complexity of a particular task can only be done on user level and we propose to do this by understanding the mental workload (MWL) generated on an operator while performing a task which requires processing a lot of information to get the task done. In this work, we have proposed an experimental setup which replicates modern day workload on doing regular day job tasks. We propose an approach to automatically evaluate the task complexity perceived by an individual by using electroencephalogram (EEG) data of a user during operation. Few crucial steps that are addressed in this work include extraction and optimization of different features and selection of relevant features for dimensionality reduction and using supervised machine learning techniques. In addition to this, performance results of the classifiers are compared using all features and also using only the selected features. From the results, it can be inferred that machine learning algorithms perform better as compared to traditional approaches for mental workload estimation.

## 1 Introduction

Industrial Sectors like aviation, military, space or transport requires continuous vigilance and multitasking by operators to perform various jobs. Hence operators are burdened by huge mental workload which causes stress and possible human errors. Thus companies and industries try to reduce the complexity of the hardware or software toolkit while increasing the intuitiveness and ease of the user interfaces of such technologies. According to [1] Mental workload is a hypothetical construct that describes the extent to which the cognitive resources required to perform a task have been actively engaged by the operator.

It not only depends on the task's characteristics but also on the person and his situation. MWL is an abstract property which comes into play during machine-human interaction and this cannot be directly observed as MWL cannot be defined inherently nor it is easy to understand the factors responsible for this. Hence it becomes difficult to build a powerful model for predicting the mental performance on a particular task. However, in the previous works, workload level has been determined through three approaches – (1) subjective measures, (2) performance-based measures and (3) physiological measures [1].

Subjective approaches depend upon the assessment of various tasks from an individual's perspective. Performance-based measures rely on the performance of the user in order to assess and determine the cognitive state [2] and physiological methods interpret the MWL using invasive, semi-invasive and non-invasive physiological techniques. Amongst all of them, physiological measurements are relatively better. These measurements aim to understand the psychological processes by their effect on the body state, rather than through perceptual ratings or performance of tasks. A number of diverse techniques are present in the literature under this category [3]; however, each one of them has some merits and demerits. So far, wide research has been carried out on human behavior and cognition using invasive techniques, such as electrocorticography (ECoG) [4] and Local Field Potentials (LFPs) [5] - [6]. In ECoG, the electrodes are placed directly on the exposed surface of the brain to record electrical activity and LFP refers to the electrical field recorded using a small-sized electrode in the extracellular space of brain tissue. These techniques involve surgery and are risky. A non-invasive BCI uses brain activities recorded from an electroencephalogram (EEG), functional Magnetic Response Imaging (fMRI) or magnetoencephalogram (MEG), etc.

Among the available non-invasive devices, EEG-based BCIs facilitate many real-time applications, as they satisfy convenience criteria (non-intrusive, non-obtrusive and simple) and effectiveness criteria (sensitive, efficient and compatible). Portable wireless and low-cost EEG devices have gained popularity for studying vigilance [7], cognitive workload [8], and motor task [9], as they not only allow for assessment of mental task directly but also because of their high temporal resolution (milliseconds). This makes EEG an appropriate tool for capturing fast and dynamically changing brain wave patterns in complex cognitive tasks. Besides, a noveler application for MWL measurement can be carried out using the wireless data acquisition systems like EEG and it can promote future research and development

Collecting EEG data is a time-consuming and tedious process. Such high dimensional EEG data processing not only delays the processing time but also affects the accuracy of the classifier. It is also important to keep in mind that the brain signal patterns observed during different times of the day and even during different sessions vary considerably [10]. This interpersonal variability of EEG signals leads to poor classifier performance. Additionally, many features, like time-domain [11], statistical [12], wavelet [13], auto-regressive coefficients [14], frequency-domain [15] features have been extracted for better understanding of the EEG signals by machine learning algorithms. But still, it is a question whether these are the best features that can be used as a state of the art methodology for feature engineering. In addition to this, real-time applications of using EEG Brain signals requires continuous acquisition, processing, and prediction. The issues mentioned above motivate us to lay down our research objectives as follows: Creation of an experimental setup consisting of tasks which relate to such real-life tasks, where the software or hardware toolkit can be improved based on feedback from the user, to make it more intuitive and easy for the user; addressing interpersonal variability; extracting a set of very relevant and important features, and increasing the prediction speed of the classifier in BCI system.

The highlights of our proposed task evaluation system are:
– Explore the viability of data acquisition using EEG in the assessment of MWL.
– Propose an experimental setup which resonates with real-life task evaluation scenarios where companies and industries want to figure out the intuitiveness of their software or hardware toolkit and how to improve them for ease in usability.
– Extract related features from the time series data of MWL
– Study the effect of feature selection and optimization on the performance of the classifier.
– Compare the experimental results, using all features and selected optimized features in terms of accuracy and time.

In this work, Emotiv Epoc+ EEG device has been used for collecting data for assessing MWL. We created an experimental setup where subjects were made to build 3-D Solidworks models in a controlled environment. The building of these models induced different workload on the subject's brain. Moreover, feature engineering was done to extract and select the most effective features for classification. Next, for the classification, machine learning models have been used to predict the type of task the subject is undergoing based on the MWL data encountered by the models. In order to make sure that the machine learning (ML) algorithms predict best results, we have used various algorithms from a pool of various optimal algorithms.

Some of the ML approaches used are:

Statistics based:
● Gaussian NB
Information based:
● XGBoost
● Decision Tree
Error based:
● SVM - (Support Vector Machines)
● MLP - (Multi-Layer Perceptron)

Similarity-based:
- KNN (k-Nearest Neighbors)

Our paper is organized as follows: Sect. 2 presents the already done work in this domain - the literature survey. Section 3 contains the methodology followed for the experiment. Section 4 explains EEG signal analysis and feature selection. Next, the experimental results obtained are presented in Sect. 5. Finally, the conclusion is outlined in Sect. 6.

## 2       Literature Survey

In [16], the intelligent systems were constructed by estimation of mental workload, using EEG features by the authors. Individual's workload is predicted using a Workload Index which uses the Gaussian Process Regression Model. Authors in [3], have also explored potentials of EEG as well as fNIRS (combined) for classifying mental workload. Rigorous efforts for classifying mental workload are being made currently. For example, in [17] EEG Features are incorporated for this task. In [2], Statistical EEG Features are extracted using multi-class linear classification and stepwise regression and classified into 4 levels. In [18], workload is classified into 7 levels using discrete wavelet transform as well as artificial neural networks. Further, in [19] identification of sensitive EEG features for detecting workload changes has been outlined. Correlation of patterns of EEG data in different tasks with workload variation has been noticed in [20]. In [21], feature selection based on Cross-Task performance and a regression model was utilized for classifying mental workload. Event Related Potentials (ERP) [22] and Fisher LDA [23] based EEG Features have been successfully used for binary MLW classification in [24]. A comparison of classical MLW assessment methods with Classification Models built via Machine Learning techniques is made in [25]. In [26], authors used a specific feature engineering method for mental workload estimation which involved extracting various important attributes of EEG Signals for classification. They extracted statistical, derivative, interval, hjorth, frequency domain, and wavelet features from the EEG Signals and classified using machine learning algorithms.

In summary, a lot of work has been done on experimenting with collection of MWL data. Especially on understanding mental workload, extraction of features and classifying for predicting mental workload on an individual. But less emphasis has been given on trying to improve the task itself and how intuitive and comfortable it pertains to the user, how can a particular hardware or software toolkit be improved in various domains so that critical tasks can be easily performed by any operator. Additionally, the previous work includes experimenting with gaming tasks or some mental task. We have tried to take a step ahead and tried to deal with real-life task experimental design, which correlates to a large extent the type of tasks an operator deals within real-life scenarios.

## 3 Experiment and Methodology

### 3.1 Subjects

Eight healthy subjects were a part of the experiment. The age of the participants varied from 21 to 24 years old. The subjects were the students of the Indian Institute of Technology, Kharagpur. The participants had a 6/6 or corrected-to-6/6 vision. Further, participants were physically and mentally fit and they did not have any neurological or psychiatric disorders. Permission was taken from each volunteer before the beginning of the experiment and choice was up to them when they wanted to perform the experiment. Additionally, instructions were given to the participants from refraining to drinking alcohol, 24 hours prior to the experiment and from consuming nicotine or caffeine 2 hours before the experiment.

### 3.2 Data Acquisition and Experiment Protocol

An Emotiv Epoc+ EEG device with a sampling rate of 128Hz has been used for data acquisition. The device had 14 channels which are AF3, F7, F3, FC5, T7, P7, O1, O2, P8, T8, FC6, F4, F8, and AF4, as well as two references (P3/P4), and it follows the international 10–20 standard for location of electrodes. Also, no mobile phones were allowed while the experiment was conducted. For data acquisition, two dedicated computer systems were utilized. One for carrying out the task for generating workload and the other for recording. Participants were instructed to avoid unnecessary physical movements in order to minimize the artifacts resulting due to muscular movement.

### 3.3 Tasks for generating workload

The participants used "Solidworks" to build 3D models, for mental workload generation. SolidWorks is a solid modeling computer-aided design (CAD) and computer-aided engineering (CAE) computer program that runs on Microsoft Windows. SolidWorks is published by Dassault Systèmes. It contains tools and features to create 3D objects from scratch and make sure that those 3D objects can be seen as real objects with a lot of information embedded into it. For our experiment, we restricted ourselves to very limited and fixed tools and features to make sure that it doesn't add a lot of complexity in our experiment design. We used the SolidWorks version 2017 in our experiment. The subjects had to build the models that had varying levels of complexity and difficulty and they had to use various tools and methods to build them. Participants declared having no history of Solidworks usage and no previous experience with BCIs. Each participant agreed with the terms and conditions of the study. Each participant was given an introductory hands-on experience with the tools and the whole software environment.

Below include the details of the experiment and the models.

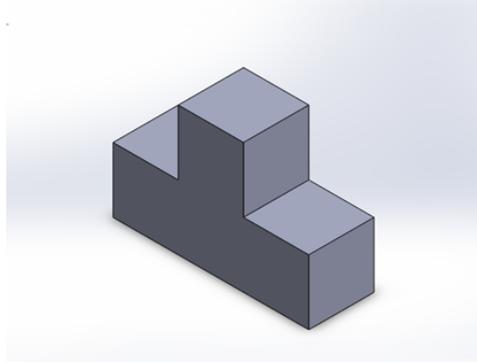

**Fig. 1.** A Solidworks model of type 1 complexity.

The first 3D model shape is shown in figure 1. This model is built in Solidworks using Sketch and Extrude tools. The time given to finish off the above-mentioned model is 50 seconds (including idle time). If the user was unable to finish the model in the given time he had to leave the system at 50 sec. This was done to make sure that each and every subject had the same time to finish off the model and so that this time limitation also induced some workload on the subject. The time allowed was assigned keeping in mind that a normal subject could finish off the model, given he has been introduced to the model-building procedure. The complexity of the model building is assumed to be of type 1.

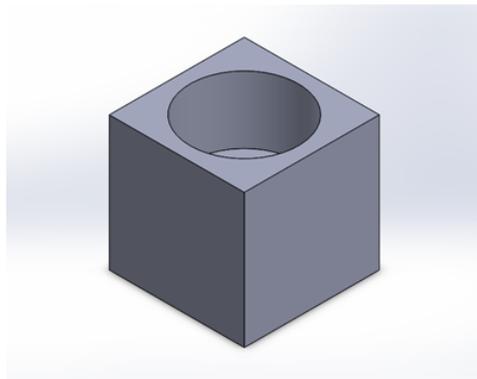

**Fig. 2.** A Solidworks model of type 2 complexity.

Figure 2 shows the second model that was given to the subjects for building on SolidWorks. The time allowed for this model was 110 seconds (including idle time). The tools required for building this model were 3 namely sketch, extrude and extrude cut. The complexity of the model building is assumed to be of type 2.

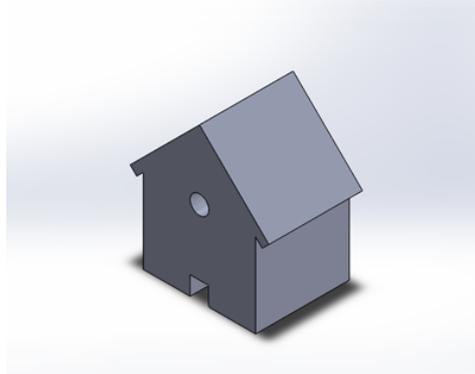

**Fig. 3.** A Solidworks model of type 3 complexity.

Figure 3 shows the third model that was given to the subjects. The time allowed for this model was 170 seconds (including idle time). The tools required for building this model were 4 namely sketch, extrude and extrude cut and mirror transform. The complexity of the model building is assumed to be of type 3.

### 3.4 Experimental Procedure

In order to provide adequate comfort to the subjects while performing the experiments, it was ensured that the experiment was done under controlled setup. The subjects build three different types of the model on Solidworks as part of the experiment. Before the experiment, we collected the personal details of the subjects which included their age, gender, medication, sleep duration, the status of mental health, education, etc. They also filled a consent form declaring their volunteering role in the experiment. The subjects had no previous experience with Solidworks or BCIs. Each participant was given an introductory hands-on experience with the tools and the whole software environment. Three different levels of models were given to each participant in any order and the experiments were conducted at three different times of the day - Morning, Afternoon and Evening. There was no association of the experimental time of the day with the level of complexity of the models. That means the subjects gave the trials in random fashion with no relation of experimental time with the model complexity. The trial order was randomly shuffled across each subject. This was done to ensure removal of any bias that may occur because of subjects building all the three models at any particular time of the day or following a particular order of model complexity while building them.

There were three levels of model complexity, type 1, type 2 and type 3. The models were shuffled before giving to the subjects for the experiment. The subjects were asked to wear the EEG device on their scalp and saline solution was applied to get a better contact of electrodes with the scalp. To minimize any of the artifacts generated, subjects were instructed to reduce the physical movements to a minimum. Subjects were also requested to refrain from excessive blinking of the eyelid. Each subject builds a model same as shown above using the software. Each subject underwent 3 trials. Each trial included the following steps-
- The first 5 seconds of procedure was quite; the subject was Idle.
- At the beginning of 6th second, the subject started building the model.
- Finally, after a given time limit and 5s, idle time after the experiment indicates the end of the trial.

**Table 1**: Details of the EEG data collection trials for different models

| One Trial | First Idle Period | Experimental time | Last Idle period |
|---|---|---|---|
| Model type 1 | 5 sec | 40 sec | 5 sec |
| Model type 2 | 5 sec | 100 sec | 5 sec |
| Model type 3 | 5 sec | 160 sec | 5 sec |

## 4 EEG Signal Analysis

Electrical signals and several undesired cerebral processes contaminate the raw EEG signals that are captured through the scalp, hence making them unfit for feature extraction. These artifacts degrade the signals of interest severely and cause alteration in the EEG signal measurements. Hence, it is important to process the EEG signals prior to the extraction of features. The pipeline includes signal pre-processing followed by extraction of features. Then we select the most important features for dimensionality reduction. Finally, the classification is done using machine learning algorithms.

### 4.1 Artifact Removal

Most of the EEG signals recorded are contaminated by artifacts and do not represent actual brain signals. Artifacts can be significantly larger than the EEG signal, and from various sources like muscle contraction or electromyogram (EMG), eye movement or electrooculogram (EOG), heart activity or electrocardiogram (ECG) and power line noise [27]. The proper study of the brain signals hence requires the removal of these artifacts. Various methods of automating the artifact removal have been proposed in the literature [28]. In this work, the recently developed technique called FORCe (Fully online and automated artifact removal) [29] for brain-computer interfacing method has been used. Following the removal of artifacts, the data is further processed.

### 4.2 Data Segmentation

From the description of the trials, the readings corresponding to the respective mental load of a particular task from all the subjects were divided accordingly. Three trials were taken for each individual, for each level of mental workload and were clubbed together. Each of these readings is further divided into 1-second epochs as shown in the figure below (see Figure 4). The sampling frequency of the device used is 128 Hz, so for an epoch of 1 second, we got 128 samples.

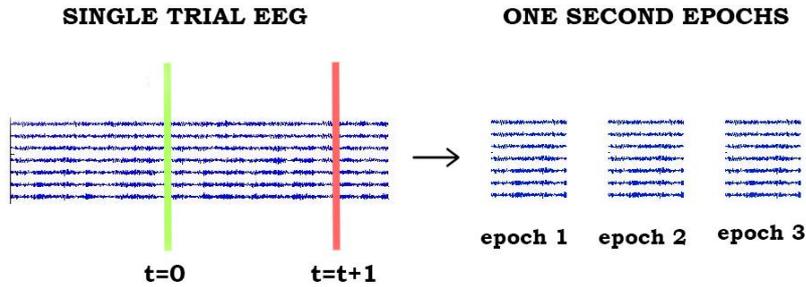

**Fig. 4.** Data segmentation of single-trial EEG into an epoch of 1 second each.

### 4.3 Feature Extraction

The filtered signals obtained for each epoch are of high dimension. It is important to reduce the complexity of such high dimension signals. We extracted the features from the processed EEG data, which gives information about distinct components of the EEG data. It is an important step, as its extraction is needed for understanding various features of the EEG data by the machine learning algorithm. We extracted a total of 52 features [26] which gives a lot of information about the type of brain waves we are dealing with and helps machine learning algorithms to better understand the data.

The details of the features extracted are given below:

**Statistical features**: The statistical features that we extracted were: median, mean, skewness, standard deviation, minimum and maximum amplitudes of the EEG data and kurtosis. These features extract the information about the distribution of amplitudes and moments of the EEG data.
**Derivative features**: Derivative features were the max and mean value of the initial two derivatives of the EEG signals.
**Interval or period features**: Analysis of EEG signals can also be done by distribution measurement of the intervals between zero and other level crossings or between maxima and minima. The features include mean and variance of the vertex to vertex amplitudes, the variance of vertex to vertex slopes, mean of the vertex to vertex times and slopes, the number of local minima and maxima, number of zero crossings of the signal, amplitude range, coefficient of variation and line length.
**Hjorth parameters**: The complexity of EEG data is indicated by Hjorth parameters. It captures the ability, mobility and the complexity of the EEG signals.

**Frequency-domain features**: These features capture the frequency information of the EEG signals; the features were extracted by applying Fast Fourier Transform (FFT) on EEG wavebands. Additionally, other important ratios of FFT were also calculated from various EEG bands.

**Wavelet features**: The wavelet transform (WT) is capable of distinguishing very small and delicate differences between time-series signals even from short signal epochs. It can easily identify highly irregular and non-stationary signals. Further, WT based methods can localize the signal components in time-frequency space in a better way than FFT analysis.

### 4.4 Feature Normalization and Selection

The features that are extracted are then normalized to bring them to a common range. This optimization helps in reduction of inter-subject variability. Here, the extracted features are mean-normalized using the following Eq.1

$$(1)$$

Overfitting and dimensionality curse can be minimized using feature selection and optimization. Features which are strongly correlated to the target variables, which in our case is the task type are selected for classification. Feature selection becomes important because it not only decreases the number of features for further processing but also increases the computation speed since the machine learning algorithm has to deal with feature space of low dimension.

Feature Selection Method Used:
- Tree-based feature selection (Extra Trees Classifier)
- Extreme Gradient Boosting (XGBoost)
- Correlation Feature Selection

**Extra Trees Classifier**: This [30] method builds multiple trees and splits nodes using random subsets of features. The splitting condition is based on impurity. This impurity for classification problems is calculated using Gini impurity / information gain (entropy), while it is variance for regression trees. So when we train the trees to predict the target variable, we can compute the contribution of each feature in predicting the correct output. It also calculates the weighted impurity of each feature for prediction which gives an insight of what features are important for classification. Feature importance using Extra trees classifier is calculated by averaging the decrease in impurity over trees.

**Extreme Gradient Boosting**: [31] It becomes relatively simpler and easier to extract the feature importance scores of each attribute by using gradient boosting. A score is provided to each feature which indicates the usefulness of the feature in decision boosted trees construction. The attributes which are used to make more key decisions with decision trees are given higher importance. This importance calculation is done for each feature in the dataset allowing the features to be compared with each other. The importance is calculated for a single decision tree by the amount that each attribute split point improves the performance measure (Gini Index), weighted by the number of observations the node is responsible for. The feature importance is then averaged across all the decision trees within the model.

**Correlation Feature Selection**: The Correlation Feature Selection (CFS) [32] evaluates the features based on the following hypothesis: "A subset of good features contains highly correlated attributes for classification, yet being uncorrelated to each other". We assume a feature to be good if it has more relevance to the prediction class, while at the same time is independent against other features for classification. So in easier terms, it can be said that we are interested in finding such features that are highly correlated to the prediction class but at the same time are not so much related to the other features used for classification. The features that had good correlation score were filtered out of the feature list for classification.

## 5  Results and Discussion

### 5.1  Selected Features and Plots

We used the three different types of feature selection techniques i.e. Trees Based Feature Selection (Extra Trees Classifier), XGBoost Feature Selection and Correlation-based Feature Selection. We were able to get the most important features for each specific feature selection and feature important techniques. Based on the results, we selected top 10 features for each selection technique. The results for each of them are as follows:

**Table 2.** Top 10 most important features from different feature selection technique.

| Extra Trees | XGBoost | Correlation |
|---|---|---|
| Wavelet Detailed STD | AR | Wavelet App. Entropy |
| Wavelet Detailed Energy | Wavelet Detailed STD | Hjorth_activity |
| Wavelet App. Entropy | Variance of V to V slope | Variance of V to V slope |
| Auto Regressor | Wavelet Appx. STD | Wavelet Detailed Energy |
| Wavelet Appx. STD | Kurtosis | Wavelet Appx. STD |
| Variance of Vertex to Ver slope | Hjorth_mobility | FFT Beta Max Power |
| Delta/ Theta | Wavelet App. Entropy | 1st Difference Max |
| Wavelet Appx. mean | Delta/ Theta | FFT Alpha Max Power |
| FFT Delta Max Power | Wavelet App. Energy | Coefficient of Variation |
| Delta/ Alpha | Wavelet Appx. Mean | FFT Theta Max Power |

Each of them had some common features and were very relevant to our experiment. It proved that some set of features were very likely to predict the outcome of the classification problem. Based on the features selected we were able to rank them in order of their relevance. Figure 5, 6 and 7 shows the scores achieved by the features on each of the technique.

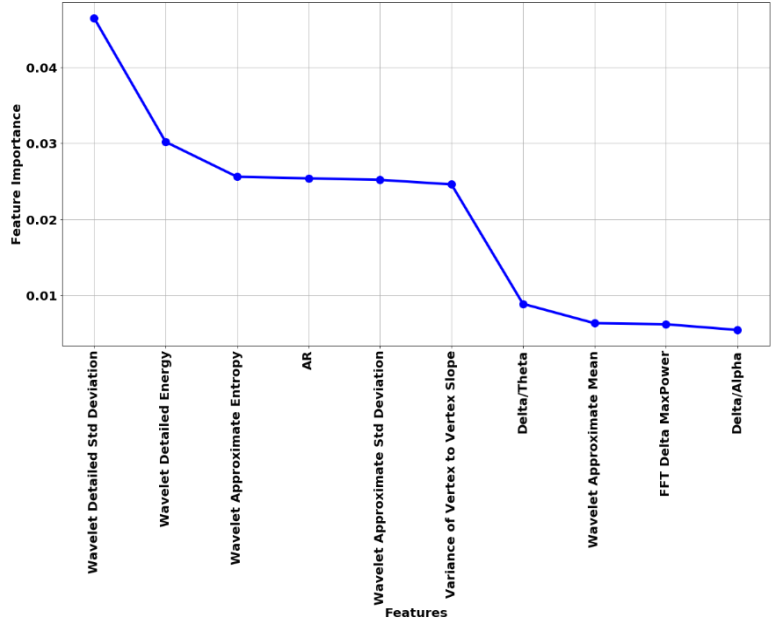

**Fig. 5.** Feature Importance Score using Extra-Trees Method

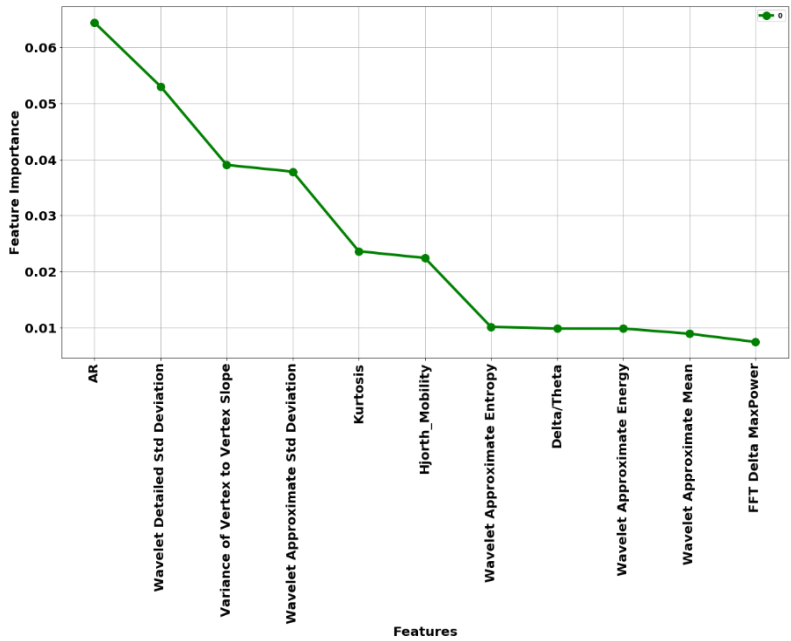

**Fig. 6.** Feature Importance Score using XGBoost Method

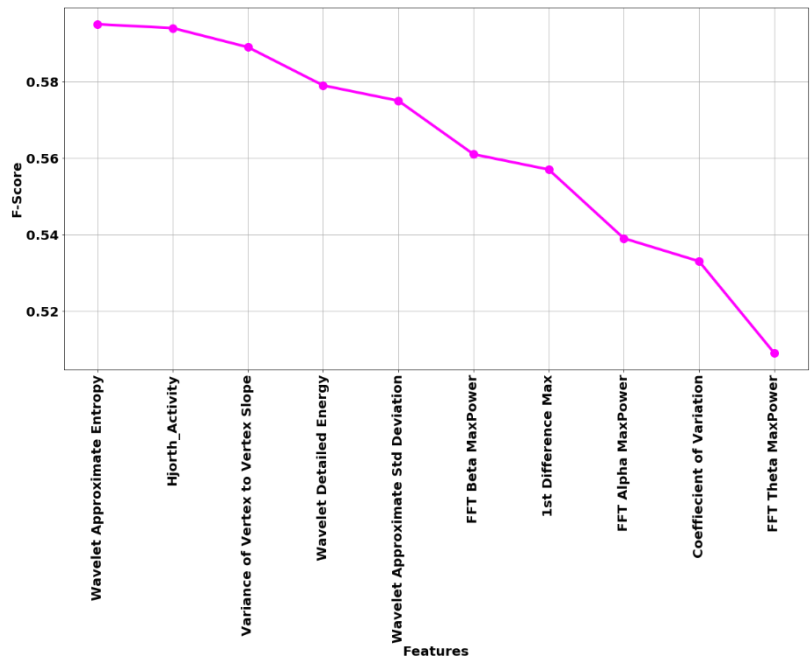

**Fig. 7.** Feature Importance using Correlation

Additionally, XGBoost provided us a unique way to see how different features were relevant in predicting the correct output. For a different number of features, the predicting power of the classifier showed the following relationship. The accuracy increased on increasing the number of features while on further increasing the features the accuracy was lost a bit. The relationship can be seen in the given scatter plot (see Figure 8).

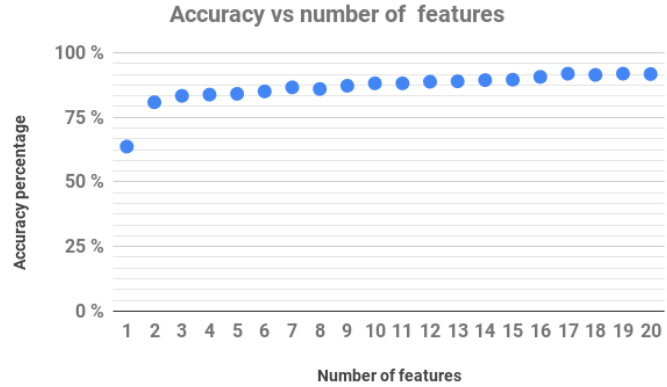

**Fig. 8.** XGBoost Accuracy v/s Number of features used for classifying the target values.

Finally, we were able to select a few set of features from the top best features from the feature selection methods. Here are the finally optimized selected features.

**Table 3.** Selected Features.

| Optimized feature set | |
|---|---|
| Kurtosis | Hjorth_Mobility |
| Wavelet Detailed Std Deviation | Wavelet Approximate Entropy |
| Variance of Vertex to Vertex Slope | Wavelet Approximate Std Deviation |
| Delta/Theta | Coefficient of Variation |
| Delta/Alpha | FFT Alpha MaxPower |
| 1st Difference Max | Wavelet Approximate Energy |
| Wavelet Detailed Energy | FFT Beta MaxPower |

Various machine learning models were used to classify the target variables.
- k-Nearest Neighbors (k-NN)
- Decision Tree Classifier
- Support Vector Machine (SVM)
- Multi-Layer Perceptron (MLP)
- GaussianNB
- XGBoost

### 5.2   Classification Accuracy Using the Classifiers

Classification of the mental workload data into different levels has been achieved by using supervised machine learning algorithms mentioned in Section 1. Four-fifths of the dataset was used to train the classifier models, and the remaining fifth of the

4dataset forming the test set. The execution of the algorithm was done using the Scikit-learn open library.

Comparisons were done in the classification of the target variable using all features and only the selected features. We found that although the selected features were comparatively less accurate but they proved to classify the target value to a great extent. Below is the table containing the overall accuracy of different machine learning models when all features were used and when only a limited number of selected features were used.

**Table 4.** Classification Accuracy and Time Taken for classification.

| Classifier | Accuracy Using all features | Time taken with all features | Accuracy using selected features | Time taken using selected features |
|---|---|---|---|---|
| Gaussian NB | 57.32% | 0.019 s | 53.74% | 0.007 s |
| Decision Tree | 79.44% | 0.121 s | 73.21% | 0.018 s |
| SVM | 82.24% | 1.041 s | 70.56% | 0.198 s |
| KNN | 86.13% | 0.456 s | 75.08% | 0.044 s |
| MLP | 84.73% | 1.937 s | 71.81% | 1.339 s |
| XGBoost | 88.01% | 0.359 s | 83.33% | 0.046 s |

**Classification without Feature Selection:**
It can be interpreted from the above results (refer Table 8) that the best accuracy is given by the XGBoost classifier. It can also be observed that the highest percentage presented by the classifier is 88.01%.

**Classification with Feature Selection:**
After feature selection and optimization, it can be observed that the classification accuracy of the classifiers decreases a bit for all the classifiers (refer Table 8) but at the same time, there is an increase in prediction speed since the time for prediction has decreased. Further, it can be noted that the XGBoost Classifier outperforms all the other classifiers Maximum accuracy obtained in the classification by XGBoost was 83.33%.

### 5.3 Confusion Matrix

A confusion matrix is used for effectively depicting the performance of the classifier (XGBoost). Each row of the matrix represents the predicted class and while each column represents the actual class (or vice -versa). We present the normalized confusion matrix of the XGBoost classifier in Figure 9 and 10. The data has been normalized by dividing the actual values by the total number of instances in the dataset to bring the range of the values between 0 and 1.

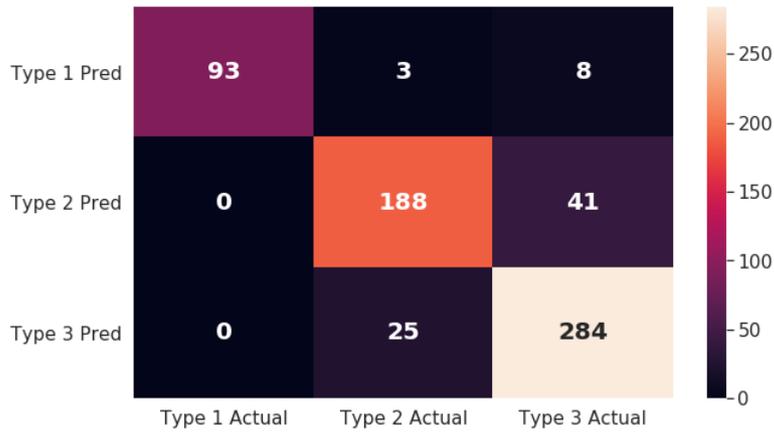

**Fig. 9.** Confusion Matrix Before Feature Selection using XGBoost Classifier.

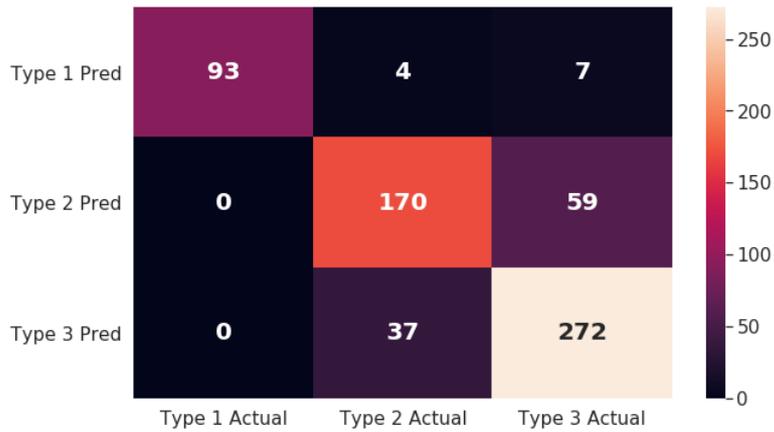

**Fig. 10.** Confusion matrix after feature selection using XGBoost Classifier.

## 6      Conclusion

One of the initial goals of the paper was to explore the feasibility of wireless data acquisition devices in task evaluation and MWL assessment. It is evident that that approach used in the work can be used to understand the complexity of a task and this information can be used by companies and industries to make their software, hardware and UI toolkits better and more intuitive for the user. In this work, we first collected the EEG signals of users in operation. Secondly, we used methods to extract features and then used only the most relevant and important features for classification of tasks. Using the feature selection method reduces computational complexity significantly which can be used for real-life prediction of the task complexity. Third, we studied the potential of ML approaches to classify the tasks done by the user into various types. From the results obtained it can be observed that XGBoost classifier gives the best accuracy in comparison to other machine learning algorithms. We hope that this study would be helpful in future to explore and devise new methods for studying and understanding the task, their complexity and the mental workload required in its operation, helping in improving the usage and user interface of various software and hardware toolkits.

1927. Benbadis, S.R.: EEG artifacts. http://emedicine.medscape.com/article/1140247-overview#a3. Accessed 14 Feb 2018.
28. Urigüen, J.A., Garcia-Zapirain, B.: EEG artifact removal-state-of-the-art and guidelines. J. Neural Eng. 12(3), 031001 (2015).
29. Daly, I., Scherer, R., Billinger, M., Müller-Putz, G.: Force: fully online and automated artifact removal for brain-computer interfacing. IEEE Trans. Neural Syst. Rehabil. Eng. 23(5), 725–736 (2015).
30. Geurts, P., Ernst, D., Wehenkel, L.: Extremely Randomized Trees, Mach Learn (2006) 63: 3. https://doi.org/10.1007/s10994-006-6226-1.
31. Machine Learning Mastery https://machinelearningmastery.com/feature-importance-and-feature-selection-with-xgboost-in-python/.
32. Doshi, M., Chaturvedi, S.K.: Correlation-based Feature Selection (CFS) Technique to predict students performance. IJCNC, Vol.6, No.3, (2014).